\pdfoutput=1
\documentclass[11pt]{article}

\usepackage[preprint]{acl}
\usepackage[utf8]{inputenc}
\usepackage[T1]{fontenc}
\usepackage{hyphenat}
\usepackage{xspace}
\usepackage{amsmath}
\usepackage{amsfonts}
\usepackage{hyperref}
\usepackage{url}
\usepackage{booktabs}
\usepackage{multirow}
\usepackage{makecell}
\usepackage{caption}
\usepackage{minibox}
\usepackage{bbm}
\usepackage{graphicx}
\usepackage{balance}
\usepackage{mathtools}
\usepackage{color}
\usepackage{marvosym}
\usepackage{ifthen}
\usepackage{textcomp}
\usepackage{enumitem}
\usepackage{verbatim}
\usepackage{algorithm}
\usepackage{numprint}
\usepackage{balance}

\usepackage{amsthm}
\theoremstyle{plain}

\newcommand{\chatoDisplayMode}[1]{#1}

\definecolor{MyRed}{rgb}{0.6,0.0,0.0} 
\definecolor{MyBlack}{rgb}{0.1,0.1,0.1} 
\newcommand{\inred}[1]{{\color{MyRed}\sf\textbf{\textsc{#1}}}}
\newcommand{\frameit}[2]{
  \begin{center}
  {\color{MyRed}
  \framebox[.9\columnwidth][l]{
    \begin{minipage}{.85\columnwidth}
    \inred{#1}: {\sf\color{MyBlack}#2}
    \end{minipage}
  }\\
  }
  \end{center}
}

\newcommand{\note}[2][]{\chatoDisplayMode{\def\@tmpsig{#1}\frameit{{\Pointinghand} Note}{#2\ifx \@tmpsig \@empty \else \mbox{ --\em #1}\fi}}}
\newcommand{\todo}[2][]{\chatoDisplayMode{\def\@tmpsig{#1}\frameit{{\Writinghand} To-do}{#2\ifx \@tmpsig \@empty \else \mbox{ --\em #1}\fi}}}

\newcommand{\abbrevStyle}[1]{#1}
\newcommand{\eg}{\abbrevStyle{e.g.}\xspace}
\newcommand{\cf}{\abbrevStyle{cf.}\xspace}

\newcommand{\vs}{\abbrevStyle{vs.}\xspace}

\newcommand{\Secref}[1]{Sec.~\ref{#1}}

\newcommand{\Tabref}[1]{Table~\ref{#1}}
\newcommand{\Figref}[1]{Fig.~\ref{#1}}

\newcommand{\Appref}[1]{Appendix~\ref{#1}}

\newcommand{\xhdr}[1]{\vspace{1.7mm}\noindent{{\bf #1.}}}

\newcommand{\textcite}[1]{\citeauthor{#1} \shortcite{#1}}

\newcommand{\hide}[1]{}

\makeatletter
\newcommand{\iffont}[2]{\ifthenelse{\equal{\f@family}{#1}}{#2}{}}
\makeatother

  \usepackage{mathptmx}

  \DeclareSymbolFont{greek}{OML}{cmm}{m}{n}
  \DeclareMathSymbol{\alpha}{\mathalpha}{greek}{"0B}
  \DeclareMathSymbol{\beta}{\mathalpha}{greek}{"0C}
  \DeclareMathSymbol{\gamma}{\mathalpha}{greek}{"0D}
  \DeclareMathSymbol{\delta}{\mathalpha}{greek}{"0E}
  \DeclareMathSymbol{\epsilon}{\mathalpha}{greek}{"0F}
  \DeclareMathSymbol{\zeta}{\mathalpha}{greek}{"10}
  \DeclareMathSymbol{\eta}{\mathalpha}{greek}{"11}
  \DeclareMathSymbol{\theta}{\mathalpha}{greek}{"12}
  \DeclareMathSymbol{\iota}{\mathalpha}{greek}{"13}
  \DeclareMathSymbol{\kappa}{\mathalpha}{greek}{"14}
  \DeclareMathSymbol{\lambda}{\mathalpha}{greek}{"15}
  \DeclareMathSymbol{\mu}{\mathalpha}{greek}{"16}
  \DeclareMathSymbol{\nu}{\mathalpha}{greek}{"17}
  \DeclareMathSymbol{\xi}{\mathalpha}{greek}{"18}
  \DeclareMathSymbol{\pi}{\mathalpha}{greek}{"19}
  \DeclareMathSymbol{\rho}{\mathalpha}{greek}{"1A}
  \DeclareMathSymbol{\sigma}{\mathalpha}{greek}{"1B}
  \DeclareMathSymbol{\tau}{\mathalpha}{greek}{"1C}
  \DeclareMathSymbol{\upsilon}{\mathalpha}{greek}{"1D}
  \DeclareMathSymbol{\phi}{\mathalpha}{greek}{"1E}
  \DeclareMathSymbol{\chi}{\mathalpha}{greek}{"1F}
  \DeclareMathSymbol{\psi}{\mathalpha}{greek}{"20}
  \DeclareMathSymbol{\omega}{\mathalpha}{greek}{"21}
  \DeclareMathSymbol{\varepsilon}{\mathalpha}{greek}{"22}
  \DeclareMathSymbol{\vartheta}{\mathalpha}{greek}{"23}
  \DeclareMathSymbol{\varpi}{\mathalpha}{greek}{"24}
  \DeclareMathSymbol{\varrho}{\mathalpha}{greek}{"25}
  \DeclareMathSymbol{\varsigma}{\mathalpha}{greek}{"26}
  \DeclareMathSymbol{\varphi}{\mathalpha}{greek}{"27}
  \DeclareSymbolFont{otone}{OT1}{cmr}{m}{n}
  \DeclareMathSymbol{\Gamma}{\mathalpha}{otone}{0}
  \DeclareMathSymbol{\Delta}{\mathalpha}{otone}{1}
  \DeclareMathSymbol{\Theta}{\mathalpha}{otone}{2}
  \DeclareMathSymbol{\Lambda}{\mathalpha}{otone}{3}
  \DeclareMathSymbol{\Xi}{\mathalpha}{otone}{4}
  \DeclareMathSymbol{\Pi}{\mathalpha}{otone}{5}
  \DeclareMathSymbol{\Sigma}{\mathalpha}{otone}{6}
  \DeclareMathSymbol{\Upsilon}{\mathalpha}{otone}{7}
  \DeclareMathSymbol{\Phi}{\mathalpha}{otone}{8}
  \DeclareMathSymbol{\Psi}{\mathalpha}{otone}{9}
  \DeclareMathSymbol{\Omega}{\mathalpha}{otone}{10}
  \DeclareSymbolFont{syms}{OML}{cmm}{m}{it}
  \DeclareMathSymbol{\partial}{\mathord}{syms}{"40}
  \DeclareMathAlphabet{\mathbold}{OML}{cmm}{b}{it}
  \DeclareSymbolFont{largesymbols}{OMX}{cmex}{m}{n}

  \DeclareMathAlphabet{\mathcal}{OMS}{cmsy}{m}{n}
\usepackage{times}
\usepackage{latexsym}
\usepackage[T1]{fontenc}
\usepackage[utf8]{inputenc}

\usepackage{microtype}

\usepackage{inconsolata}

\usepackage{graphicx}

\title{Edisum: Summarizing and Explaining Wikipedia Edits at Scale}

\author{Marija \v{S}akota  \\
  EPFL \\
  {\fontsize{11pt}{12pt}\selectfont\texttt{marija.sakota@epfl.ch}} \\\And
  Isaac Johnson \\
  Wikimedia Foundation \\
  {\fontsize{11pt}{12pt}\selectfont\texttt{isaac@wikimedia.org}} \\\And
  Guosheng Feng \\
  EPFL \\
  {\fontsize{11pt}{12pt}\selectfont\texttt{guosheng.feng@epfl.ch}} \\\And
  Robert West \\
  EPFL \\
  {\fontsize{11pt}{12pt}\selectfont\texttt{robert.west@epfl.ch}}
}
\newcommand{\model}{Edisum}

\begin{document}
\maketitle
\begin{abstract}
% An \textit{edit summary} is a succinct comment written by a Wikipedia editor explaining the nature of, and reasons for, an edit to a Wikipedia page. Edit summaries are crucial for maintaining the encyclopedia: they are the first thing seen by content moderators and they help them decide whether to accept or reject an edit. Additionally, edit summaries constitute a valuable data source for researchers. 
% Unfortunately, as we show, for many edits, summaries are either missing or incomplete. To overcome this problem and help editors write useful edit summaries, we propose a model for recommending edit summaries generated by a language model trained to produce good edit summaries given the representation of an edit diff. This is a challenging task for multiple reasons, including mixed-quality training data, the need to understand not only what was changed in the article but also why it was changed, and efficiency requirements imposed by the scale of Wikipedia. We address these challenges by curating a mix of human and synthetically generated training data and fine-tuning a generative language model sufficiently small to be used on Wikipedia at scale. Our model performs on par with human editors. Commercial large language models are able to solve this task better than human editors, but are not well suited for Wikipedia, while open-source ones fail on this task. More broadly, this paper showcases how language modeling technology can be used to support humans in maintaining one of the largest and most visible projects on the Web.

An \textit{edit summary} is a succinct comment written by a Wikipedia editor explaining the nature of, and reasons for, an edit to a Wikipedia page. Edit summaries are crucial for maintaining the encyclopedia: they are the first thing seen by content moderators and they help them decide whether to accept or reject an edit. Additionally, edit summaries constitute a valuable data source for researchers. 
Unfortunately, as we show, for many edits, summaries are either missing or incomplete. To overcome this problem and help editors write useful edit summaries, we propose a model for recommending edit summaries generated by a language model trained to produce good edit summaries given the representation of an edit diff. To overcome the challenges of mixed-quality training data and efficiency requirements imposed by the scale of Wikipedia, we fine-tune a small generative language model on a curated mix of human and synthetic data. Our model performs on par with human editors. Commercial large language models are able to solve this task better than human editors, but are not well suited for Wikipedia, while open-source ones fail on this task. More broadly, we showcase how language modeling technology can be used to support humans in maintaining one of the largest and most visible projects on the Web.

\end{abstract}

\section{Introduction}

\begin{figure}[t]
    \centering
    \includegraphics[width=\columnwidth]{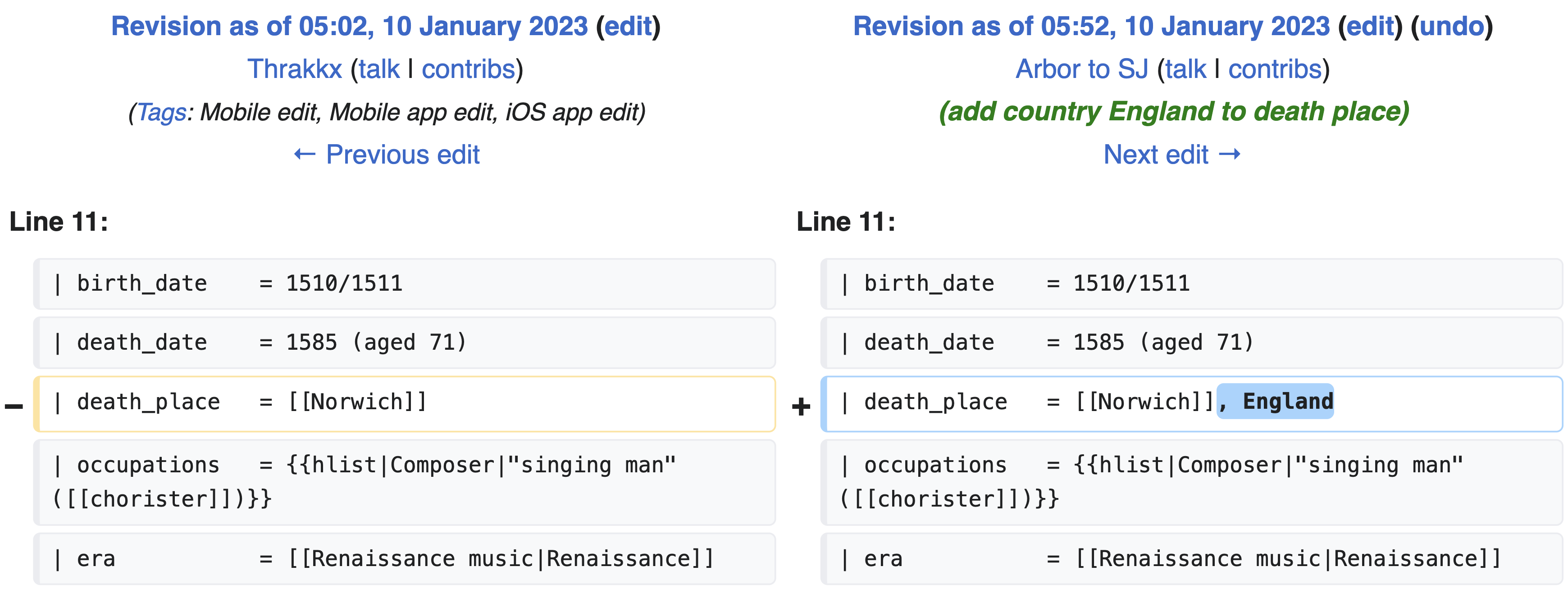}
    \vspace{-2mm}
    \caption{An example of an edit diff. The + and -- signs denote the text that was added and removed, respectively. The edit summary is the text in green in the screenshot.}
    \label{fig:edit_example}
    \vspace{-2mm}
\end{figure}

Wikipedia is the largest online encyclopedia, housing 60 million articles in over 300 languages, with the English Wikipedia alone featuring 6.7 million entries. It is edited collaboratively, meaning that anyone can be an editor to most of the articles, resulting in massive numbers of edits performed continuously; \eg, on English Wikipedia alone, over 3 million edits are performed each month \citep{num_edits}.
When performing an edit, the editor can leave an \emph{edit summary} (%also known as \emph{edit comment};
example in \Figref{fig:edit_example}), a short comment explaining the content of the edit and, sometimes, a reason why the edit was performed. It is often the first source of information about an edit that editors see when browsing edit histories for content moderation or other purposes and is an opportunity for an editor to justify their changes. %An editor might then choose to view the edit diff to see if the changes match the summary and are reasonable.

Edit summaries are also valuable to researchers. They provide important insights into editor roles and actions on Wikipedia \citep{10.1145/1718918.1718941,arazy2016turbulent,wattenberg2007visualizing}. They are used for building datasets for various purposes, such as the detection of low-quality Wikipedia content \citep{asthana2021automatically} or detecting conflicts \citep{Sumi_2011}. Edits and edit summaries have also been used to build datasets for iterative text generation, due to their incremental nature \citep{schick2022peer, faltings-etal-2021-text}.

Despite being a valuable asset, edit summaries have a number of drawbacks that prevent them from being used more efficiently. Many users leave them blank when performing an edit. Even when provided, summaries can be misleading---and not necessarily deliberately (as opposed to vandalism). Some editors also use canned edit summaries \citep{canned_summaries}, to quickly insert commonly used summaries in the current Wikipedia space. For instance, these can be edit summaries such as ``Added links'' or ``Fixed typo''. They are not intentionally misleading, but frequently do not reflect the content of the edit precisely.
Although it is hard to mitigate the effects of vandalism on edit summaries, our analyses show that a large fraction of edits would benefit from a more specific, tailored summary. This is currently an unexplored area within research, with no previous attempts to automatically generate Wikipedia edit summaries. Given their performance on text generating tasks, generative language models arise as a promising solution.

Generating Wikipedia edit summaries is a tricky problem for several reasons. Although blank edit summaries are easy to detect, there are no established heuristics for determining whether an edit summary is a good description of its edit or not. This can lead to mixed-quality data for training a model, and, consequently, poor performance at deployment time. Furthermore, edit summaries should ideally explain why the edit was performed, along with what was changed, which often requires external context.
%which requires inferring information that is not explicitly present in the text that was edited.
From an engineering perspective, it is also not trivial to design an appropriate prompt for this task, in particular because most generative models work with a small context size.
% which is another constraint that has to be taken into account.
% Finally, a generative model used for this task at scale cannot be a large language model (LLM). Even though LLMs are promising candidates for automatic edit summary generation in theory, deploying an LLM-based system would require immense expenses, especially for platforms like Wikipedia, where vast numbers of users would rely on the model every day. Furthermore, use of open-source technology is a guiding principle for Wikimedia \citep{open_source}.
Finally, even though LLMs are promising candidates for automatic edit summary generation in theory, platforms like Wikipedia often have guiding principles which limit them to the usage of open-source technology \citep{open_source}, which limits their use of commercial LLMs such as OpenAI models.

In this paper, we perform a detailed qualitative analysis of edit summaries, uncovering some of the drawbacks of human-written ones. We show that this task can be %, at a small scale, 
solved with LLMs. Based on these results, we carefully select a high-quality subset of edits and edit summaries. For a subset of the edits lacking summaries, we generate edit summaries using an LLM. Due to efficiency and input-size constraints, we then fine-tune a range of smaller generative language models with longer context size, built on LongT5 \citep{guo2022longt5}, which we call \emph{\model}. We use mix of editor-provided and synthetic data, using a representation of edit diffs as inputs. This approach balances providing sufficient context for most edits while remaining scalable for platforms like Wikipedia. %We build on a model, LongT5 \citep{guo2022longt5}, that allows for a longer context size. For the input, we use a representation of an edit diff. For the majority of edits, this is a good choice that provides a model with enough context, while still being short enough to be used with a subset of generative models. This small model can effectively be used on a large scale required by Wikipedia.

\xhdr{Results} We compare our solution to two baselines, human editors\footnote{We envision that our model could provide a recommended edit summary to human editors, simplifying the process of writing it and encouraging uptake.} and the far more resource-heavy LLMs (GPT-4, GPT-3.5 and Llama 3 8B), via both automatic and human evaluation. Our results indicate that commercial LLMs (GPT-4 and GPT-3.5) outperform both open-source LLM (Llama 3 8B) and human editors, while \model\ trained on synthetic data matches human editors' performance, offering an ideal solution for a large-scale application on Wikipedia. %Our results indicate that while GPT-4 is effective (but expensive), \model\ trained on synthetic data matches human editors' performance, offering an ideal solution for a large-scale application on Wikipedia.

\xhdr{Contributions} In short, our contributions are the following:

(i) We perform a comprehensive qualitative analysis of the existing Wikipedia edit summaries, which shows that many existing edit summaries have hard-to-detect flaws.

(ii) We show that edit summary generation is solvable using high-performance LLMs.

(iii) We show that \textit{\model}, which is, to the best of our knowledge, the first solution to
automate the generation of highly-contextual Wikipedia edit summaries at large scale,
achieves performance similar to the human editors

(iv) We release the dataset consisting of cleaned edit summaries and synthetically generated data for future research. The code can be found at %\url{[anonymous]}. 
\url{https://github.com/epfl-dlab/edisum}.

\section{Related work}

%\subsection{Wikipedia edit summaries}
\xhdr{Wikipedia edit summaries}
Edit summaries play an important role on Wikipedia in helping patrollers quickly monitor edits for vandalism or otherwise problematic edits \citep{wikipedia_edit_summary}. They are simpler and easier to scan than the edit diffs, and thus are important for enabling fast patrolling of content on Wikipedia~\citep{morgan2019patrolling}. Despite this, we are not aware of work that focuses on helping editors to improve edit summaries.

One related task that was studied more is git commit message generation. While this area is well studies, with many rule-based approaches \cite{rule-based-1, rule-based-2}, retrieval approaches \cite{retrieval-based}, learning-based approaches \cite{commitbert, learning-based-1, learning-based-2}, or even an attempt to solve the task with LLMs \cite{lopes2024commitmessagesagelarge}, the difference lies in the data. Code and textual data have many differences, with the most notable one for our problem being the lack of highly structured text that exists in the code. Wikipedia edits and edit summaries have also higher variety in the topics they cover, as well as style they are written with.

Edit summaries have been used extensively, however, to understand and model behavior on Wikipedia. Researchers who utilize edit summaries occasionally comment on anecdotal patterns in usage, but descriptive statistics are minimal. \citet{panciera2009wikipedians} describe the usage of links to Wikipedia policy pages in edit summaries, showing that the likelihood of invoking a policy increases with editor experience. \citet{wattenberg2007visualizing} convert edit summaries into colors to visualize how different editors approach tasks on Wikipedia and \citet{geiger2011trace} describe the importance of edit summaries in tracing activity on Wikipedia for understanding bots and vandalism, while \citet{stvilia2008information} point out that edit summaries are often blank or misleading, rendering them less useful. Multiple works \citep{yang2017identifying,pavalanathan2018mind,asthana2021automatically} construct datasets of edits for training models by filtering edits based on certain keywords in the edit summaries. Notably, \citet{yang2017identifying} classify edits based on their intention, including labels commonly found in edit summaries, such as ``clarification''. In contrast to their multi-label classification method, we opt for a more flexible generative language model approach. %They treat this challenge as multi-label classification, however, as opposed to our focus on the more flexible generative language model approach.

%\subsection{Synthetic data generation}
\xhdr{Synthetic data generation}
\label{sec:sgd_background}
Early approaches using generative models to produce synthetic data focused on finetuning a pretrained model which is then used as a generator \citep{anaby2020not, papanikolaou2020dare, mohapatra2020simulated, kumar2020data}. This requires an existing dataset for finetuning the generator.
Recently, the focus has shifted on unsupervised methods for synthetic data generation using pretrained language models (PLMs). These methods do not require lengthy and expensive labeling. %besides a few demonstrations sent to the PLM. 
One such example is the work by \citet{DBLP:journals/corr/abs-2109-09193}, in which they generate synthetic labels by using only unlabeled examples sent to the LLM. There have been several attempts to generate data for different natural language processing (NLP) tasks by carefully designing prompts to the PLMs. This includes work by \citet{zerogen} and \citet{zerogen+} in which they evaluate this procedure on text classification, question answering, and natural language inference tasks. Similarly, \citet{https://doi.org/10.48550/arxiv.2202.04538} do this for GLUE~\citep{wang2018glue} tasks. There have been successful attempts to use synthetic data generated in this way for intent classification~\citep{intent}, and question answering~\citep{https://doi.org/10.48550/arxiv.2212.08635}.

There are also examples of synthetic data generation for more tailored purposes. \citet{https://doi.org/10.48550/arxiv.2302.00618} use the synthetic data as demonstrations to improve the propmting of LLMs. Additionally, synthetic data has been used to solve tasks that LLMs cannot directly solve, such as closed information extraction \citep{josifoski2023exploiting}.
Our task is not a standard NLP task, such as text classification or summarization, but can still be seen as a text generation task. As such, it is likely that LLMs can solve it with careful prompting, enabling synthetic data generation for training a more efficient system suitable for large-scale use.  %This suggests the possibility of leveraging LLMs to generate synthetic data for training a more efficient system suitable for large-scale use.%This opens the opportunity for us to use synthetically generated data by LLMs for training a smaller, more efficient system to be used for large scale application.
\section{Qualitative analysis of Wikipedia edit summaries}
\label{sec:qual_analysis}
\begin{table*}
\centering
\resizebox{\textwidth}{!}{
\setlength{\tabcolsep}{5pt}
\begin{tabular}{@{}l|p{0.25\textwidth}p{0.27\textwidth}p{0.35\textwidth}p{0.27\textwidth}p{0.27\textwidth}p{0.27\textwidth}}
\toprule
\textbf{Metric} & Summary (what) & Explain (why) & Misleading & Inappropriate & Generate-able (what) & Generate-able (why) \\
\midrule

\textbf{Description} &
  Attempts to describe what the edit did. For example, ``added links'' &
  Attempts to describe why the edit was made. For example, ``Edited for brevity and easier reading''. &
  Overly vague or misleading per English Wikipedia guidance. For example, ``updated'' without explaining what was updated is too vague. &
  Could be perceived as inappropriate or uncivil per English Wikipedia guidance. &
  Could a language model feasibly describe the ``what'' of this edit based solely on the edit diff. &
  Could a language model feasibly describe the ``why'' of this edit based solely on the edit diff.\\

\midrule

\textbf{\% Agreement}     & 0.89 & 0.8 & 0.77& 0.98 & 0.97 & 0.8 \\
\textbf{Cohen's Kappa}     & 0.65 & 0.57 & 0.50 & -0.01 & 0.39 & 0.32 \\
\textbf{Overall (n=100)}     & 0.75 - 0.86 & 0.26 - 0.46 & 0.23 - 0.46 & 0.00 - 0.02 & 0.96 - 0.99 & 0.08 - 0.28 \\
\midrule
\textbf{IP editors (n=25)}     & 0.76 - 0.88 & 0.20 - 0.44 & 0.40 - 0.64 & 0.00 - 0.08 & 0.92 - 0.96 & 0.04 - 0.16 \\
\textbf{Newcomers (n=25)}     & 0.76 - 0.84 & 0.36 - 0.48 & 0.24 - 0.52& 0.00 - 0.00 & 0.92 - 1.00 & 0.12 - 0.20 \\
\textbf{Mid-experienced (n=25)}     & 0.76 - 0.88 & 0.28 - 0.52 & 0.16 - 0.36 & 0.00 - 0.00 & 1.00 - 1.00 & 0.08 - 0.28 \\
\textbf{Experienced (n=25)}     & 0.72 - 0.84 & 0.20 - 0.40 & 0.12 - 0.32 & 0.00 - 0.00 & 1.00 - 1.00 & 0.08 - 0.48 \\
\bottomrule
\end{tabular}
}
\vspace{-2mm}
\caption{Statistics on agreement for qualitative coding for each facet and the proportion of how many edit summaries met each criteria. Ranges are a lower bound (both of the coders marked an edit) and an upper bound (at least one of the coders marked an edit). The majority of summaries are expressing only what was done in the edit, which we also expect a language model to do. A significant portion of edits is of low quality, i.e., misleading.}
\label{tab:qual_coding}
\vspace{-5mm}
\end{table*}

%Ranges are a lower bound (neither of the coders marked an edit) and an upper bound (both of the coders marked an edit) but not standard confidence intervals and no comparison tests are conducted between groups.

Given the dearth of data on the nature and quality of edit summaries on Wikipedia, we perform qualitative coding to guide our modeling decisions. Specifically, we analyze a sample of 100 random edits made in August 2023 to English Wikipedia stratified among a diverse set of editor expertise levels. Two of the authors each coded all 100 summaries and we report the results in Table~\ref{tab:qual_coding}. Since there were only two coders, we report the range for each category instead of the majority label. The lower bound indicates both annotators marked the category, and the upper bound indicates at least one did. Edit summaries were coded by following criteria set by the English Wikipedia community \citep{wikipedia_edit_summary} (see \Tabref{tab:qual_coding}).
For more details on the annotation process, see \Appref{appendix:qual_analysis}.

%Ranges are a lower bound (neither of the coders marked an edit) and an upper bound (both of the coders marked an edit) but not standard confidence intervals and no comparison tests are conducted between groups.

Overall, we see a relatively high annotator agreement. Lower Cohen's kappa for some categories indicates that these judgements can be difficult and highly subjective.
The vast majority ($\sim$80\%) of current edit summaries focus on ``what'' of the edit, with only 30--40\% addressing the ``why''. This aligns with the raters' judgement of what a language model can generate from the edit diff alone (see columns ''Generate-able (what)`` and ''Generate-able (why)`` in \Tabref{tab:qual_coding}). 
Accurately describing the ``why'' requires external context that the model lacks, such as information about sources added or world events.
%Accurately describing the ``why'' would require external context that a model does not have access to, such as information about a source being added or events happening in the world.

A sizeable minority ($\sim$35\%) of edit summaries were labeled as ``misleading'', generally due to overly vague summaries or summaries that only mention part of the edit. This makes training on this data challenging. Almost no edit summaries are inappropriate, likely because highly inappropriate edit summaries would be deleted \citep{revision_deletion} by administrators and not appear in our dataset. This suggests that it is unlikely for a model trained on edit summaries to learn to suggest inappropriate summaries and thus we do no further filtering of summaries for inappropriate language.
\section{Method}
\label{sec:method}

\subsection{Synthetic data generation}

From the analysis in \Secref{sec:qual_analysis}, we notice there is a considerable number of lower quality edits, which are not easily detectable. At the same time, as LLMs perform well for a wide variety of tasks, often in a few-shot setting, we expect them to generate a good quality edit summary after some prompt tuning for majority of the edits, including what was done, but also why the edit was performed when obvious from the context. Our initial exploration on GPT-4 and GPT-3.5 confirms these assumptions. Our idea is not to just prompt LLMs to solve the task, but to rather generate synthetic data which will be used to tune a more efficient model.%At the same time, we expect a language model to be able to generate what was done in the majority of the edits and why the edit was performed when obvious from the context.
%As LLMs perform well for a wide variety of tasks, often in a few-shot setting, we expect them to generate a good quality edit summary after some prompt tuning. Our initial exploration on GPT-4 and GPT-3.5 confirms these assumptions. 

\xhdr{LLM}
\label{sec:llm}
After experimenting with available OpenAI models \citep{openai_models}, we opt for gpt-3.5-turbo model with 4k token context as a good compromise between price and quality of the results. This model is optimized for the dialogue setting. We prompt it by sending the explanation of what an edit summary is as a system prompt, while the demonstrations are presented as alternating dialogue turns by the user (edit diff) and the model (edit summary).

Generating useful synthetic training data requires an LLM that can already solve the task of automated edit summary generation---the very task we set out to solve---, which might seem to defeat the purpose of this paper.
We hence emphasize that commercial LLMs are not well suited for this task, as they do not follow the open-source guidelines set by Wikipedia \citep{open_source}. In addition, we envision this model as an assistant to the editors, meaning that it should run virtually in real-time. Given the low number of GPUs Wikipedia has access to \cite{gpu_access}, ideally, our model should be fairly small to fulfil the real-time constraints.
%We hence emphasize that LLMs do not fit the bill for the kind of solution we seek, due to their size and the prohibitive maintenance and deployment costs, especially in a high-throughput regime as faced by Wikipedia, where hundreds of thousands of edit summaries need to be generated every day. For instance, using 100 random edits to calculate the average prompt length, running GPT-4 with 8k-token context, just for English Wikipedia, would cost above \$1300 per day, assuming the current rate of 3 million edits per month \citep{model_pricing}. These are considerable expenses and do not scale to over 300 languages of Wikipedia. Using GPT-3.5 with 4k-token context, the cost would be around \$25 per day. Despite the lower cost of this model, the OpenAI models are not open-sourced, so using them would be against the principles of Wikipedia.

\xhdr{Prompt construction}
\label{sec:prompt_construction}
We settle on the five-shot setting, instructing the LLM to only explain what was done in the edit, as the reason why the edit was performed is often too difficult to infer from the context. Nonetheless, we observe that LLMs often generate the reason organically where it is appropriate.\footnote{For instance, for the edit \url{https://en.wikipedia.org/w/index.php?diff=1172890678} GPT-4 will generate ``Removed unnecessary quotation marks around the name Claudia.'', hinting that the edit was performed because the quotation marks were unnecessary.} The examples of edits with good summaries, represented with the edit diff between the revision immediately before \vs\ after the edit, are used as demonstrations (see \Figref{fig:edit_example}).
The edit diff is much shorter than the full revisions, which makes it easier to fit our prompt into the length constraints imposed by the LLM. Additionally, the edit diff provides rich information about what was performed during the edit, omitting a large amount of text that was irrelevant for the edit. For more details on the prompt tuning and quality check of generated data, see \Appref{appendix:synth_data_process}.

%%%%%%%%%%%%%% data cleaning

\subsection{Data cleaning and collection}
\label{sec:data_cleaning}
We filter Wikipedia data for training the models with two aspects in mind. First, \emph{\textbf{edit summaries for certain types of edits are trivial}}. For example, HotCat, a tool that many editors use to change categories on a page, automatically generates reasonable summaries via heuristics (e.g. ``added \href{https://en.wikipedia.org/wiki/Category:Shoegazing_musical_groups}{Category:Shoegazing musical groups} using \href{https://en.wikipedia.org/wiki/Wikipedia:HotCat}{HotCat}'',\footnote{\url{https://en.wikipedia.org/w/index.php?diff=1033805631}}).
Based on this, we focus on edits altering the text of the article, where heuristics struggle and a language model would be well-suited. Second, \emph{\textbf{existing edit summaries are of mixed quality}}, which is reflected in the qualitative coding described in Section~\ref{sec:qual_analysis}.
This is most salient in IP editors, and, to a lesser degree, new editors.
In this context, we exclude the following edits:

(i) \textbf{Edits which did not change}, insert, or remove at least one \textbf{sentence }in the article.

(ii) \textbf{Edits with auto-generated summaries} by Mediawiki software \citep{automatic_summaries}.

(iii) \textbf{Edits made by bots}, which often have very good edit summaries, but it is not useful to have a language model learn edit summaries that have already been hard-coded into a bot.

(iv) \textbf{Reverted edits}, as many of them are vandalism and unlikely to have a useful summary.

(v) \textbf{Edits that made the revert} to previous edits, as these often talk about reason why the revert was performed. These reasons are usually external and are not easy to infer from the edit diff, and thus are difficult to be generated by a language model.

(vi) \textbf{Edits with blank edit summaries}, as all edits should have a basic edit summary. Many edits have an indicator of which section of the article they affected, which we removed from all edits as well, so it does not affect our checking of whether the summary is blank.

We also annotate the edit summaries with the various metadata (e.g. length) to enable further filtering or balancing of our edit summary sample (see \Appref{appendix:data annotation}).

%%%%%%%%%%%%% model

\subsection{Model}
Since 4.6\% of our data requires input size longer than 512 tokens used by standard small generative models \citep{chung2022scaling, lewis2019bart}, as the model to finetune, we use LongT5 \citep{guo2022longt5}, which has the ability to work with longer context windows. We denote each finetuned model as \model[$S\%$], where $S$ is a percentage of synthetic data in the training set. We intentionally use a very small model, because of limitations of Wikipedia's infrastructure. In particular, Wikipedia does not have access to many GPUs on which we could deploy big models \citep{gpu_access}, meaning that we have to focus on the ones that can run effectively on CPUs. Note that this task requires a model running virtually in real-time, as edit summaries should be created when edit is performed, and cannot be precalculated to decrease the latency. Models of similar size have already been successfully implemented in Wikipedia applications. %While most of our data has input shorter than the usual 512 tokens that most of the smaller generative models work with \citep{chung2022scaling, lewis2019bart}, 4.6\% of it requires a longer context size. 
For details on implementation, see \Appref{appendix:model_implementation}.

Because this is an unexplored area, with no previous attempts to automatize the generation of Wikipedia edit summaries, there is no apparent baseline to compare against. We thus directly compare our method to the actual ground-truth data: edit summaries written by Wikipedia editors. In addition to that, we evaluate how close our model is to LLMs. We evaluate GPT-4 and GPT-3.5, which we used to generate synthetic data. Additionally, we evaluate an open-source alternative LLM of a reasonable size, Llama 3 8B \cite{llama3modelcard}. We ran all LLMs on 500 randomly chosen edits from the test data and they were all prompted with the same prompt used for synthetic data generation, and with the generation parameters from \Tabref{tab:generation_params}.
\section{Experimental setup}
%%%%%%%%%%%%%%%%%%%%%%%%%%%%%%%%%%%%%%%%%%%%%%%%%%%%%%%%%

\subsection{Data}
\label{sec:data}

We use edits made in August 2023 to articles on English Wikipedia. This includes over 500K edits without a summary, from which we randomly take 100k edits to generate synthetic data. 
After the initial filtering from \Secref{sec:data_cleaning}, we are left with $\sim$600K edits. We additionally limit the data by filtering edits with summaries longer than 200 and shorter than 5 characters. We leave out edits from the editors who have made less than 30 edits and keep at most 3 edits with the same summary to enforce diversity. This leaves us with $\sim$127K samples.
For experiments, we combine data obtained in both ways: from existing Wikipedia edit summaries, and by generating synthetic data. We use in total 100K samples for training, and 10K for validation. The rest is used for testing ($\sim$17K samples).

We run experiments with 5 different proportions of synthetic data in the training set (0\%, 25\%, 50\%, 75\%, 100\%), by choosing the synthetic and human editor's data randomly from the collected datasets. 
As input to the model, we use the edit diff between the two revisions of the article in question to keep the input short while preserving the most important information. We extract the difference and represent the input in the same manner as in \Secref{sec:prompt_construction}. To separate the text from the old and the new revision, we use <old\_text> and <new\_text> prefixes. Each sentence is separated by <sent\_sep> prefix. We filter out data points with inputs longer than 1,024 tokens for convenience (only 2.3\% of our data is longer than that). As the output, we use the (human- or synthetically generated) edit summary. For an example of the constructed input, see \Appref{appendix:model_input}.

\subsection{Evaluation}
We perform a twofold evaluation:
(1) a cheap and fast-to-conduct automatic evaluation in which we compare auto-generated summaries to the human-written ones (ground truth); and % that uses human-written edit summaries as the ground truth against which auto-generated summaries are compared; and
(2) an expensive and slower-to-conduct human evaluation, where human raters compare auto-generated to human-written edit summaries.
In the former case, the best a model can do is reproduce a human-written summary, whereas in the latter case, a model can in principle outperform humans on this task.

\xhdr{Automatic evaluation}
For automatic evaluation, we use MoverScore~\citep{zhao-etal-2019-moverscore}, designed for measuring the semantic similarity between two texts.
It takes values from 0 to 1 (larger is better), and correlates better with human judgement than token\hyp matching metrics such as BLEU \citep{10.3115/1073083.1073135} or ROUGE \citep{lin-2004-rouge}.
This is especially important in settings similar to ours, where many good outputs with different phrasing may be equally appropriate.
To evaluate a single \model\ model or a single LLM, for each edit, we take edit diff, generate the automatic summary with it, and calculate the MoverScore by comparing it to the existing summary. We obtain the measure of quality for the current \model\ model by averaging this measure over the whole dataset. For a reference, we also provide ROUGE and BERT scores obtained in the same way in~\Appref{appendix:automatic_eval}.

\xhdr{Human evaluation}
\label{sec:human_eval}
Although data cleaning increases the overall quality of the edit summaries we consider, some of them are still misleading or incorrect, as we do not have a good heuristic to detect this. Yet, MoverScores are obtained by comparing to those existing edit summaries, which can result in scores that have little to no meaning. To surpass this limitation, we perform a human evaluation.
We compare our best-performing model according to the MoverScore (\cf\ \Secref{sec:auto_eval_res}), \model[100\%] (trained fully on synthetic data), with summaries written by editors and GPT-4 (highest performing model from \Secref{sec:auto_eval_res}). To inspect the effect of synthetic data on training, we also evaluate \model[0\%], trained only on existing data.
%We compare two of our models with the summaries written by human editors and GPT-4 (highest performing model from automatic evaluation). We choose the best-performing \model\ model according to the MoverScore (\cf\ \Secref{sec:auto_eval_res}), trained with 100\% synthetic data (\model[100\%]). To inspect the effect of synthetic data on training, we also evaluate the version with no synthetic data (\model[0\%]).

%\xhdr{Annotation task}
We randomly select 100 samples from the testing dataset to perform this evaluation, from which we discard one sample without a good edit summary option. Each sample corresponds to a Wikipedia edit, and is associated with a web page of the edit diff between two revisions\footnote{\eg, \url{https://en.wikipedia.org/w/index.php?title=Albert_Einstein&diff=prev&oldid=1177682587}; see \Figref{fig:edit_example} for a visual example}. For each sample, annotators are presented with four edit summaries in random order, to prevent bias: ground truth, \model[100\%], \model[0\%], and GPT-4 summary. They are asked to choose the best and the worst summary, because it is often difficult to rank all four summaries, as some of them are very similar or convey the exact same information, in which case the preference would only come down to the style of the summary (see \Tabref{tab:summaries_examples} for examples). The task can be seen as ranking with ties, where the two summaries that were not chosen as neither are tied for the second place. %For more details on the task, see \Appref{appendix:human_eval_setup}.
%\xhdr{Annotators}
Since this is not a simple task, to ensure high-quality results, instead of relying on the crowdsourcing platforms, we recruited 3 MSc students to perform the annotation. Conflicts were resolved by one of the authors of the paper. To measure the agreement between the annotators, we report Kendall rank correlation coefficient (Kendall's $\tau$) between each pair of them. As our annotation task can be seen as a ranking task, we choose this as a suitable measure. For more details on annotation task, see \Appref{appendix:human_eval_setup}.

\section{Results}

\begin{figure}[htbp]
    \centering
    \includegraphics[width=0.95\columnwidth]{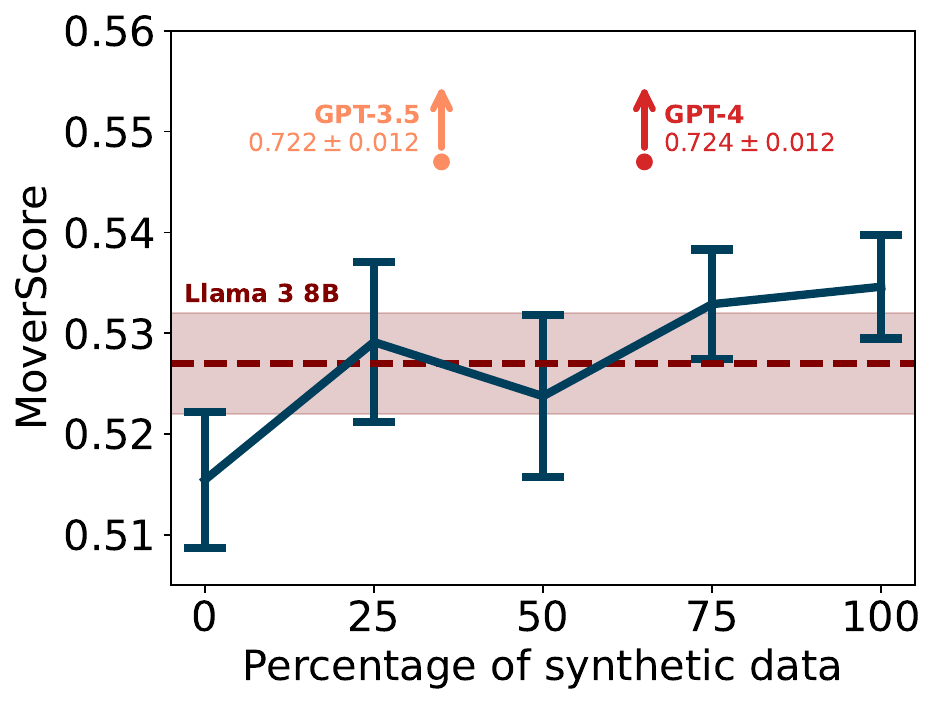}
    \caption{Results of \model\ evaluation with MoverScore. Error bars are 95\% confidence intervals (CIs). GPT-4 and GPT-3.5 perform better than \model, with the average MoverScore of 0.724 and 0.722, respectively. We do not show the performance of GPT-4 and GPT-3.5 credibly on y-axis for convenience, as their performance is susbstantially higher than for the other models. Note that both of these are shown as a dot on the plot, as there is no notion of the percentage of synthetic data in the training set for these models.}
    \vspace{-2mm}
    \label{fig:auto_results}
    \vspace{-2mm}
\end{figure}
\subsection{Automatic evaluation}
\label{sec:auto_eval_res}

In \Figref{fig:auto_results}, we present the results of automatic evaluation. %\model\ models and a subset of testing dataset using GPT-4 and GPT-3.5.
Performance of all \model\ models is decent, according to the MoverScore. \model[0\%] performs worse than the models with some fraction of synthetic data, in particular \model[75\%] and \model[100\%], for which this difference is also statistically significant. This confirms our assumption that synthetic data is a useful asset when tackling the task of edit summary generation.
%While one might be surprised that a fully-synthetic training set results in a model that matches the existing data better than the one trained on the existing data itself, this is not unexpected.
One might be surprised that a fully-synthetic training set results in higher score when comparing to the existing data than the training set with only existing data, but this is not unexpected.
Existing data has more structural variety and features various Wikipedia tags, which can be hard for a language model to pick up. Synthetic data might not have the same surface form as the existing data, but it expresses the key information about the edit while maintaining simpler structure, making it easier to train on.%On the other hand, even though synthetic data might not always have the same surface form as the existing data, it still maintains the essence of the information contained in it, while maintaining a simpler structure across all the edit summaries.

When it comes to LLMs, as anticipated, the results show that commercial ones effectively solve this task, achieving scores higher than any of the \model\ models. The difference between GPT-4 and GPT-3.5 is small. We suspect this happens because we did not tune the prompt or generation parameters specifically to GPT-4. Further tuning can only improve the results, confirming the usefulness of these LLMs. On the other hand, the open-source LLM, Llama 3 8B, underperforms even when compared to the finetuned \model\ models. %Nevertheless, we expect that with tuning of the prompt and parameters, this result could only be better, further confirming our statement that these models can generate useful edit summaries. 
%These models, however, are not suitable for use in a setting like Wikipedia, where large number of editors are performing thousands of edits every day (See \Secref{sec:llm} for cost estimation and limitations). This is why it is essential to have a smaller model that can do a decent enough job. For trickier edits, we could still reach out to an LLM. This approach would tremendously lower the costs of running such a system.
Given the limitations Wikipedia has on using only open-source software and their low performance on this task, as well as the need for this model to be fast and efficient, it is essential to have a smaller model that can do a decent job. This approach would also lower the costs of running such a system. For similar applications without such constraints, GPT-4 would be a reasonable option.

%%%%%%%%%%%%%%%%%%%%%%%%%%%%%%%%%%%%%%%%%%%%%

\subsection{Human evaluation}

\begin{figure}
    \centering
    \includegraphics[width=\columnwidth]{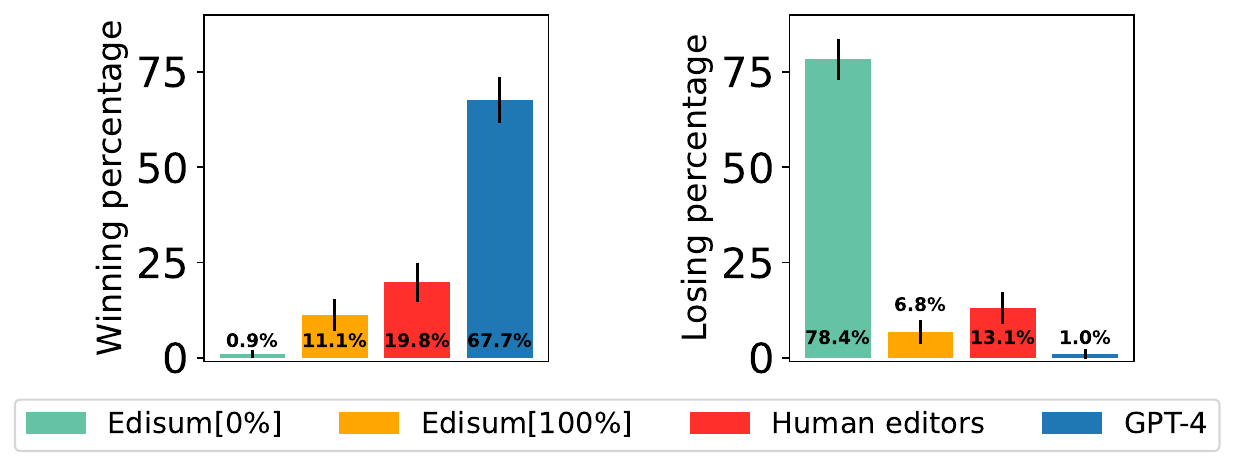}
    \caption{Results of human evaluation. \emph{Left:} \% of time summaries from each method are chosen as the best. \emph{Right:} \% of time summaries from each method are chosen as the worst. Error bars are 95\% confidence intervals (CIs).}
    \vspace{-2mm}
    \label{fig:human_eval_results}
    \vspace{-1mm}
\end{figure}
\begin{figure}
    \centering
    \includegraphics[width=0.85\columnwidth]{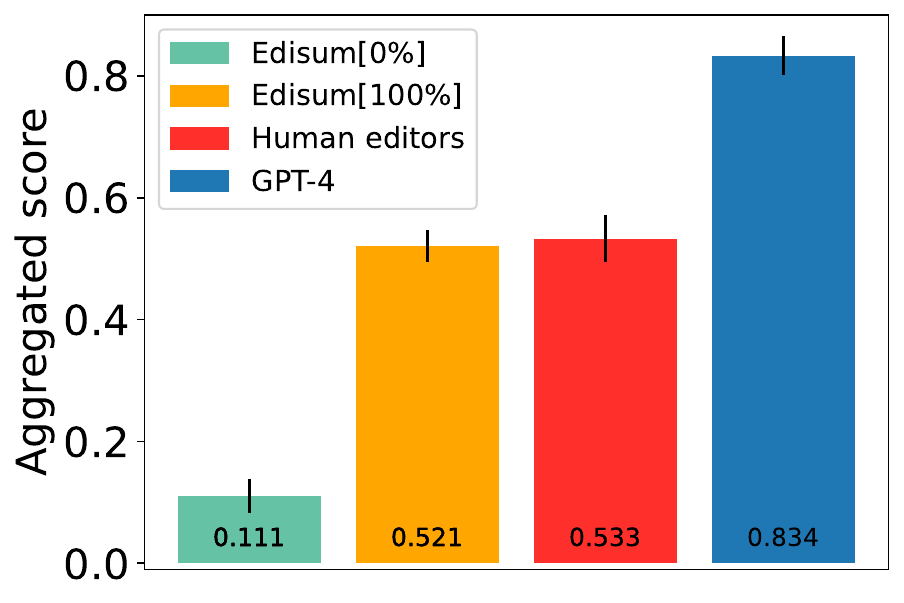}
    \caption{Average aggregated scores of human evaluation. Each method was scored with 1 point for winning, 0 points for losing, and 0.5 for neither winning nor losing. Error bars are 95\% confidence intervals (CIs).}
    \vspace{-2mm}
    \label{fig:agg_scores}
    \vspace{-1mm}
\end{figure}
\begin{table*}[t]
\centering
\resizebox{\textwidth}{!}{
\setlength{\tabcolsep}{3pt}

\begin{tabular}{p{25mm}cccccc|ccc}
\toprule
& \multicolumn{6}{c}{\textbf{What}} & \multicolumn{3}{c}{\textbf{Why}} \\

     \textbf{Method} &   Correct & No change & Not specific & Unclear & Unexhaustive &   Unrelated &   Correct &   Incorrect &   Missing \\
\midrule
\multicolumn{7}{l}{\emph{\textbf{Human editors}}} \\
\hspace{2mm} Win & 0.65 \scriptsize± 0.07 & 0.00 \scriptsize± 0.00 & 0.00 \scriptsize± 0.00 & 0.05 \scriptsize± 0.03 & 0.25 \scriptsize± 0.06 & 0.05 \scriptsize± 0.03 & 0.70 \scriptsize± 0.06 & 0.00 \scriptsize± 0.00 & 0.30 \scriptsize± 0.06 \\
\hspace{2mm} Lose & 0.15 \scriptsize± 0.06 & 0.00 \scriptsize± 0.00 & 0.15 \scriptsize± 0.05 & 0.39 \scriptsize± 0.06 & 0.23 \scriptsize± 0.06 & 0.08 \scriptsize± 0.03 & 0.23 \scriptsize± 0.06 & 0.08 \scriptsize± 0.03 & 0.69 \scriptsize± 0.06 \\
\hspace{2mm} Neither & 0.59 \scriptsize± 0.06 & 0.00 \scriptsize± 0.00 & 0.35 \scriptsize± 0.07 & 0.06 \scriptsize± 0.03 & 0.00 \scriptsize± 0.00 & 0.00 \scriptsize± 0.00 & 0.53 \scriptsize± 0.07 & 0.00 \scriptsize± 0.00 & 0.47 \scriptsize± 0.07 \\
\multicolumn{7}{l}{\emph{\textbf{GPT-4}}} \\
\hspace{2mm} Win & 0.92 \scriptsize± 0.03 & 0.00 \scriptsize± 0.00 & 0.04 \scriptsize± 0.03 & 0.00 \scriptsize± 0.00 & 0.04 \scriptsize± 0.03 & 0.00 \scriptsize± 0.00 & 0.36 \scriptsize± 0.07 & 0.00 \scriptsize± 0.00 & 0.64 \scriptsize± 0.06 \\
\hspace{2mm} Neither & 0.40 \scriptsize± 0.07 & 0.00 \scriptsize± 0.00 & 0.04 \scriptsize± 0.03 & 0.00 \scriptsize± 0.00 & 0.32 \scriptsize± 0.06 & 0.24 \scriptsize± 0.06 & 0.48 \scriptsize± 0.06 & 0.12 \scriptsize± 0.04 & 0.40 \scriptsize± 0.06 \\
\multicolumn{7}{l}{\emph{\textbf{Edisum[100\%]}}} \\
\hspace{2mm} Win & 0.63 \scriptsize± 0.07 & 0.00 \scriptsize± 0.00 & 0.00 \scriptsize± 0.00 & 0.00 \scriptsize± 0.00 & 0.28 \scriptsize± 0.06 & 0.09 \scriptsize± 0.04 & 0.45 \scriptsize± 0.06 & 0.00 \scriptsize± 0.00 & 0.54 \scriptsize± 0.07 \\
\hspace{2mm} Lose & 0.00 \scriptsize± 0.00 & 0.14 \scriptsize± 0.05 & 0.00 \scriptsize± 0.00 & 0.00 \scriptsize± 0.00 & 0.14 \scriptsize± 0.05 & 0.71 \scriptsize± 0.06 & 0.14 \scriptsize± 0.05 & 0.00 \scriptsize± 0.00 & 0.86 \scriptsize± 0.04 \\
\hspace{2mm} Neither & 0.34 \scriptsize± 0.06 & 0.00 \scriptsize± 0.00 & 0.09 \scriptsize± 0.04 & 0.06 \scriptsize± 0.03 & 0.19 \scriptsize± 0.05 & 0.31 \scriptsize± 0.06 & 0.28 \scriptsize± 0.06 & 0.16 \scriptsize± 0.05 & 0.56 \scriptsize± 0.07 \\
\bottomrule
\end{tabular}
}
\vspace{-2mm}
\caption{Error analysis results.}
\label{tab:error_analysis}
\vspace{-3mm}
\end{table*}

Recall that in the human evaluation, for each edit, raters were asked to pick the best and the worst one out of four summaries, each generated by one of four methods: human editors, \model[0\%], \model[100\%], and GPT-4.
The four candidate summaries for each edit were evaluated by three independent raters. Inter-rater agreement, measured in terms of Kendall's $\tau$, was 0.588, 0.556, and 0.562 for the three pairs of raters, indicating a relatively strong positive agreement among the raters.
In \Figref{fig:human_eval_results}, we report the wins and losses separately. The left and right subfigure show the percentage of edits for which each method was chosen as the best and worst, respectively. GPT-4 is chosen the most often as the best model and the least often as the worst, while \model[0\%] is the opposite. More importantly, the human editors and \model[100\%] are tied on a middle ground, with the editors being chosen slightly more often as the best, but also as the worst, compared to \model[100\%].

Since we did not let annotators compare the two middle options, to confirm our analysis, we fit a Plackett-Luce model, a generalization of the Bradley-Terry model \cite{BT}, intended to model ranking data (with the ability to handle ties, as in our setting). Briefly, this model assumes that there is a latent utility parameter associated with each option (in our case each method) and infers a maximum likelihood estimate from the empirically observed rankings (one ranking per human labeled data point). The higher the utility, the more preferred the option is. The results are presented in \Appref{appendix:plackett-luce}, and they show no statistically significant difference between \model[100\%] and editors. Moveover, we consider specifically those rankings where \model[100\%] and human data are not tied (46 out of 99 samples). \model\ ranked higher 22 out of the 46 samples (vs. 24 for editors). This difference is not statistically significant (we ran a binomial test, with p-val = 0.883).

To compute a single performance score per method, we awarded a method a score of 1 if it was chosen as the best one, 0 as the worst one, and 0.5 if it was not chosen as neither. In \Figref{fig:agg_scores}, we report the average score obtained by each method. In line with \Figref{fig:human_eval_results}, we observe that
GPT-4 scores best and \model[0\%] scores worst, while the average scores of \model[100\%] and editors are nearly identical and not statistically significantly different.

These results indicate that \model[100\%] performs equally well as human editors, but with less variance: it achieves similar average ranking scores as the human editors (\Figref{fig:agg_scores}), while taking extreme positions less often than it (\Figref{fig:human_eval_results}).
Overall, results confirm the conclusions from the automatic evaluation. 
The positive effects of synthetic training data are even more evident here. %, with the difference between using only synthetic \vs\ only human training data being even bigger.
Similarly, GPT-4 is again observed to generate edit summaries of the highest quality. However, as noted in \Secref{sec:llm} and \Secref{sec:auto_eval_res}, running such a system on a daily basis on a platform as big as Wikipedia for all the edits would not be feasible today.
Our ``distilled'' \model[100\%] model, which aims to mimic GPT's high-quality summaries, offers a fertile middle ground, performing as well as humans while being much smaller and cheaper to run.

\subsection{Error analysis}

To further examine the difference in performance between GPT-4 and other methods, we manually inspect 150 edit summaries from two perspectives: ``why'' (description of why the edit was performed) and ``what'' (description of what was done). The samples were chosen to cover all the cases (win, lose or neither) for all three methods. For details on the annotation procedure and taxonomy, see \Appref{appendix:error_annotation}. The results are presented in \Tabref{tab:error_analysis}.
%Under ``why'' category, we choose between 3 options: missing, correct, and incorrect. For ``what'', we use the taxonomy from \Tabref{tab:error_taxonomy}

When observing ``why'' meta-category, we notice, as expected, that human written summaries express the correct reason why the edit was performed more often than the ones generated by GPT-4 or \model. However, both methods frequently express the, usually correct, reason. This reflects the edits for which reason can be inferred from the context. %This reflects the edits for which the LLMs are able to infer why the edit was performed organically.
When it comes to the results for ``what'' category, the performance gap between GPT-4 and other methods is still visible. 
Specifically for \model, we can attribute the drop in performance to its size. \model\ is a very small model ($\sim$220M parameters), incapable of fully capturing patterns present in more complex tasks, like edit summary generation. %To confirm these assumptions, we ran an ablation study by finetuning a range of different-sized models. For results, see \Appref{appendix:ablation_size}.
%Specifically for \model, we can attribute this drop in performance due to a much smaller size of the model (223M parameters), making it much harder for a model to capture the variety present in the data, confirming that this is not a simple task, as it requires a good understanding of the context provided as the input.
The distribution of errors for GPT-4 and \model\ for summaries that were not chosen as neither the best or worst is similar, with the most errors being unrelated or unexhaustive summaries. On the other hand, human editor's summaries from this category, as well as the ones chosen as the worst, tend to be less specific or unclear. Summaries that won were most often not exhaustive enough. Overall, while it should be used with caution due to a portion of unrelated summaries, the analysis confirms that \model\ is a useful option that can aid editors in writing edit summaries.
\section{Conclusion}
In this paper, we investigate the quality of Wikipedia edit summaries, i.e., short comments that editors write when performing changes in Wikipedia. These summaries serve a wide range of purposes in Wikipedia, but also for general research community. We find that a non-negligible number of them is of bad quality or missing. At the same time, we show that GPT-4 is able to solve this task better than human editors. To assist editors, we train a small language model that can, unlike GPT-4, effectively generate edit summaries on a large scale while matching the performance of human editors.

\section*{Limitations}
While the overall results show that \model\ performs on par with human editors, there is still a space for improvement given that GPT-4 still outperforms our model. Additionally, the nature of the errors produced by \model\ and human editors is not the same. We leave it to the future research to explore the possibility of bridging the gap between a small generative model and a high-performing LLM and the impact different errors could have.

Our experiments show that models trained on synthetic data outperform those trained on existing edit summaries on Wikipedia, but this approach likely has limitations in learning editor community norms such as common abbreviations.

Additionally, our dataset might suffer from lack of diversity, and hence, our models might fail on more exotic edits. We limited our training samples to edit summaries by editors with at least 30 edits based on our qualitative analysis of existing edit summaries, but future work could explore additional strategies for producing a high-quality, diverse dataset of existing edit summaries. \cite{kocetkov2022stack} found significant improvements from applying near de-duplication to their code dataset and we suspect that many edits are quite similar with minor differences and a similar pipeline might bring improvements to this task as well.

% \marija{TODO}
% \begin{itemize}
%     \item While the results are overall on par with human editors, there is still space for improvement (as GPT-4 still is the best)
%     \item While the overall results are matching human editors, the nature of errors are not the same - this might need some further investigation
%     \item while we find that models trained on synthetic data outperform those trained on existing edit summaries on Wikipedia, this approach likely has limitations in learning editor community norms such as common abbreviations
%     \item We limited our training samples to edit summaries by editors with at least 30 edits based on our qualitative analysis of existing edit summaries, but future work could explore additional strategies for producing a high-quality, diverse dataset of existing edit summaries. \cite{kocetkov2022stack} found significant improvements from applying near de-duplication to their code dataset and we suspect that many edits are quite similar with minor differences and a similar pipeline might bring improvements to this task as well.
% \end{itemize}

\bibliography{custom}

\appendix
\section{Qualitative analysis annotation process}
\label{appendix:qual_analysis}

\xhdr{Data sample}
The qualitative analysis was performed on 100 samples, as annotation of edits is a lengthy process. To ensure a diverse enough group of edits, we stratify the sample based on the experience of editors. More precisely, we divide editors in four categories: IP editors (anonymous editors), newcomers (editors with < 10 edits), mid-experienced editors (10 - 1000 edits), and experienced editors (1000+ edits). We exclude edits by bots,\footnote{Anecdotally, many bots actually have very good edit summaries generated by their code but also their edits are usually straightforward to describe.} edits without summaries, and revert-related edits to focus on good-faith contributions to English Wikipedia. This means that our sample likely lacks highly-inappropriate comments, as they would be removed by the editors. Notably, 46\% of the edits from August 2023 do not have a summary. The proportion varies greatly by editor type: 74\% of edits for IP editors, 46\% of edits for newcomers, 58\% for medium-experienced editors, and 38\% of edits for experienced editors. This highlights the value of better support for generating edit summaries.

\xhdr{Annotation}
Annotation was done by the two authors of the paper. A discussion was held after the first ten summaries to ensure there was agreement on the codebook before completing the sample.

\section{Data annotation}
\label{appendix:data annotation}

We annotated the cleaned edit data as follows:

(i) \textbf{Frequency with which the edit summary appears} in our dataset. This enables us to sample the dataset to be more diverse. To check frequency, we lower-case all characters and replace any links with a generic link character before calculating frequency.

(ii) \textbf{Editor's edit count.} This enables us to up-sample edits from more experienced editors, who are expected to be more likely to write correct edit summaries.

(iii) \textbf{Summary length}. While very short summaries are okay (\eg, ``ce'' is often used to stand for ``copy-edit'' and indicate small grammar or spelling changes), summaries are limited to 500 characters and the English Wikipedia community suggests to avoid unnecessarily long summaries.\footnote{\url{https://en.wikipedia.org/wiki/Help:Edit_summary}}

(iv) \textbf{User frequency} in the dataset. A small number of editors make a large proportion of edits on Wikipedia, and while they may write reasonable edit summaries, we want to learn from a diverse sample of Wikipedians.

(v) \textbf{Semi-automated edits.} If an edit is made through a tool that enables very quick editing or has several preset edit summaries, we flag this, as these edit summaries are unlikely to be strongly contextualized to the specific edit.

\section{Synthetic data generation process}
\label{appendix:synth_data_process}

\begin{figure*}[ht]
    \centering
    \includegraphics[width=0.93\textwidth]{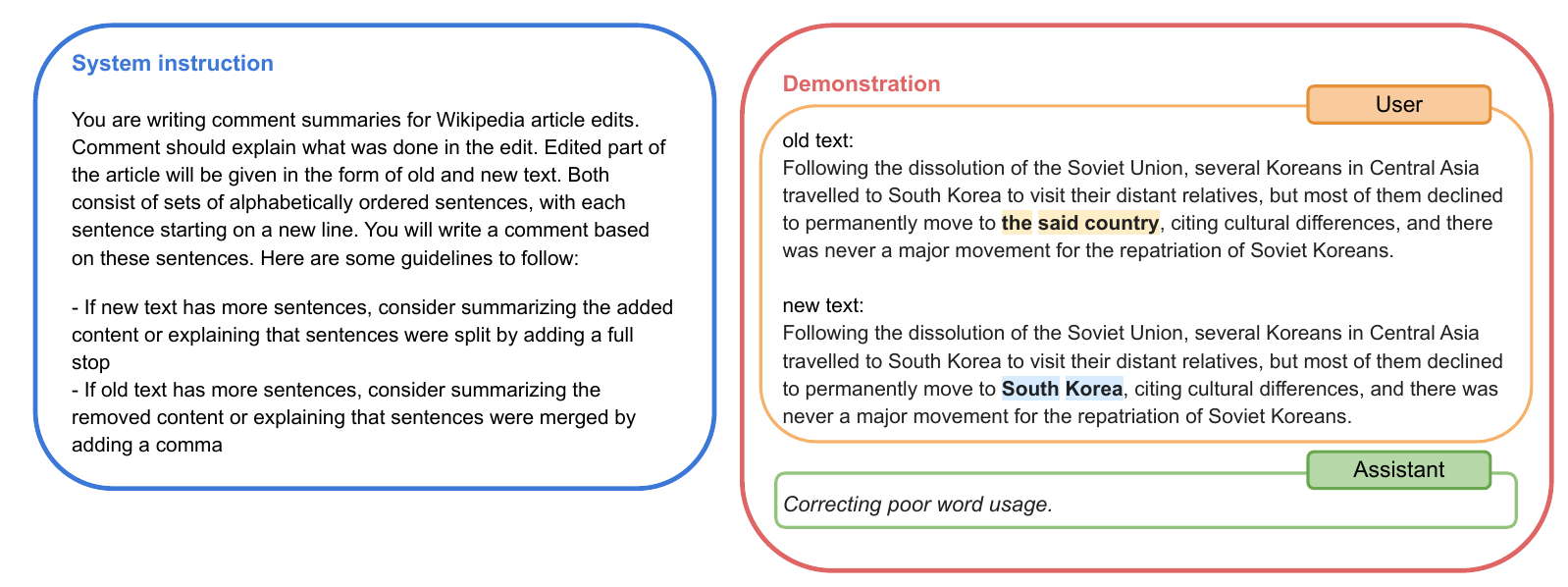}
    \caption{System instruction and the example of a demonstration used for synthetic data generation.}
    \label{fig:prompt_example}
\end{figure*}
\xhdr{Prompt choice}
%%%%%%%%%%%%%%%%%%%%%%%%%%%%%
Experimentation process for choosing the prompt is done of 10 samples of edit diffs. For each one of them, we generate an edit summary with different prompts, and after manual inspection, we settle on the prompt that is used for synthetic data generation. We experiment with different instructions and different numbers of demonstrations, as well as their content.

For the instruction, as already mentioned in \Secref{sec:prompt_construction}, we only focused on asking the LLM to explain what was performed in the edit. We also explained the format of the edit summary and the input, and gave a few guidelines to follow. For the full instructions, see \Figref{fig:prompt_example}.

For the demonstrations, as explained in \Secref{sec:prompt_construction} we provide the LLM with the edit diff between the two revisions immediately before \vs\ immediately after the edit. We extract this diff using the mwedittypes\footnote{\url{https://pypi.org/project/mwedittypes/}} library. From the output of this library, we can extract sentences that were added and removed in the editing process. We group all the removed sentences into ``old text'' and all the added sentences into ``new text''. On 100 randomly chosen and manually inspected edit diff outputs using this library, in 4 cases, these sentences are not ordered by the way they are appearing in the revision of the Wikipedia page. Because of that, we order the sentences in ``old text'' and ``new text'' alphabetically, to avoid confusion. We then represent the diff by concatenating both of those, separating them by stating ``old text:'' and ``new text:'' before each group. For an example of a demonstration, see \Figref{fig:prompt_example}.

When choosing the demonstrations, we make sure to include both longer and shorter edits in terms of content, and also edits with summaries of various length. We include both demonstrations for edits that add and remove content. We tried out different numbers of demonstrations ranging from 2 to 10 (2, 3, 5, 6, and 10 demonstrations), using same demonstrations for each edit summary generated. We settled on five demonstrations, which we found to provide us with sufficient information to generate high-quality data, while keeping the length of the input shorter, and consequently, the cost of the generation process smaller. The same five demonstrations are used for all the generated samples.

% \usepackage{booktabs}
\begin{table}[h]
\centering
\resizebox{\columnwidth}{!}{
\setlength{\tabcolsep}{5pt}
\begin{tabular}{@{}l|cccccccc@{}}
\toprule
\textbf{parameter} & max\_tokens & temperature & top-p & frequency\_penalty & presence\_penalty & stop & n & best\_of \\
\midrule
\textbf{value} & 1000 & 0 & 1 & 0.2 & 0 & "\textbackslash{}n" & 1 & 1 \\
\bottomrule
\end{tabular}
}
\caption{Generation parameters used with gpt-3.5-turbo to generate synthetic data. These parameters were also used when testing GPT-4 and GPT-3.5 performance on the testing dataset.}
\label{tab:generation_params}
\end{table}

% \begin{table}
% \centering
% \resizebox{\columnwidth}{!}{
% \setlength{\tabcolsep}{5pt}
% \begin{tabular}{@{}l|cccccccc@{}}
% \toprule
% \textbf{parameter} & max\_tokens & temperature & top-p & frequency\_penalty & presence\_penalty & stop                & n & best\_of \\
% \midrule

% \textbf{value} & 1000 & 0 & 1 & 0.2 & 0 & "\textbackslash{}n" & 1 & 1  \\

% \bottomrule
% \end{tabular}
% }
% \caption{Generation parameters used with gpt-3.5-turbo to generate synthetic data.}
% \label{tab:generation_params}
% \end{table}
\xhdr{Generation parameters and process}
We decide to only work with edits that have textual changes and exclude the ones with changes in the Wiki markup, such as category modifications or templates. We do this because this is where we expect the language model to give us the biggest gains, as this is where the biggest variety of different edits are performed. We experimented with different generation parameters for the OpenAI models. In particular, we tried out different values of temperature, top-$p$,
and frequency penalty. We make the decision on the best parameters manually, based on the same 10 samples we used for prompt construction. The set of parameters for the best-performing setup is displayed in \Tabref{tab:generation_params}.

\xhdr{Quality check}
To verify that the quality of generated synthetic data is satisfying, we perform a quality check on 100 random edits, by comparing the data generated by GPT-3.5 with the existing one. We find that synthetic data has satisfying quality more often (87\% vs. 78\% of the time). On top of that, in 30\% of the cases in which both summaries were seen as suitable, the generated one was chosen as better 30\% of the time (vs 4\% for the existing summaries).

\section{Model input}
\label{appendix:model_input}

\begin{figure*}[ht]
    \centering
    \includegraphics[width=\textwidth]{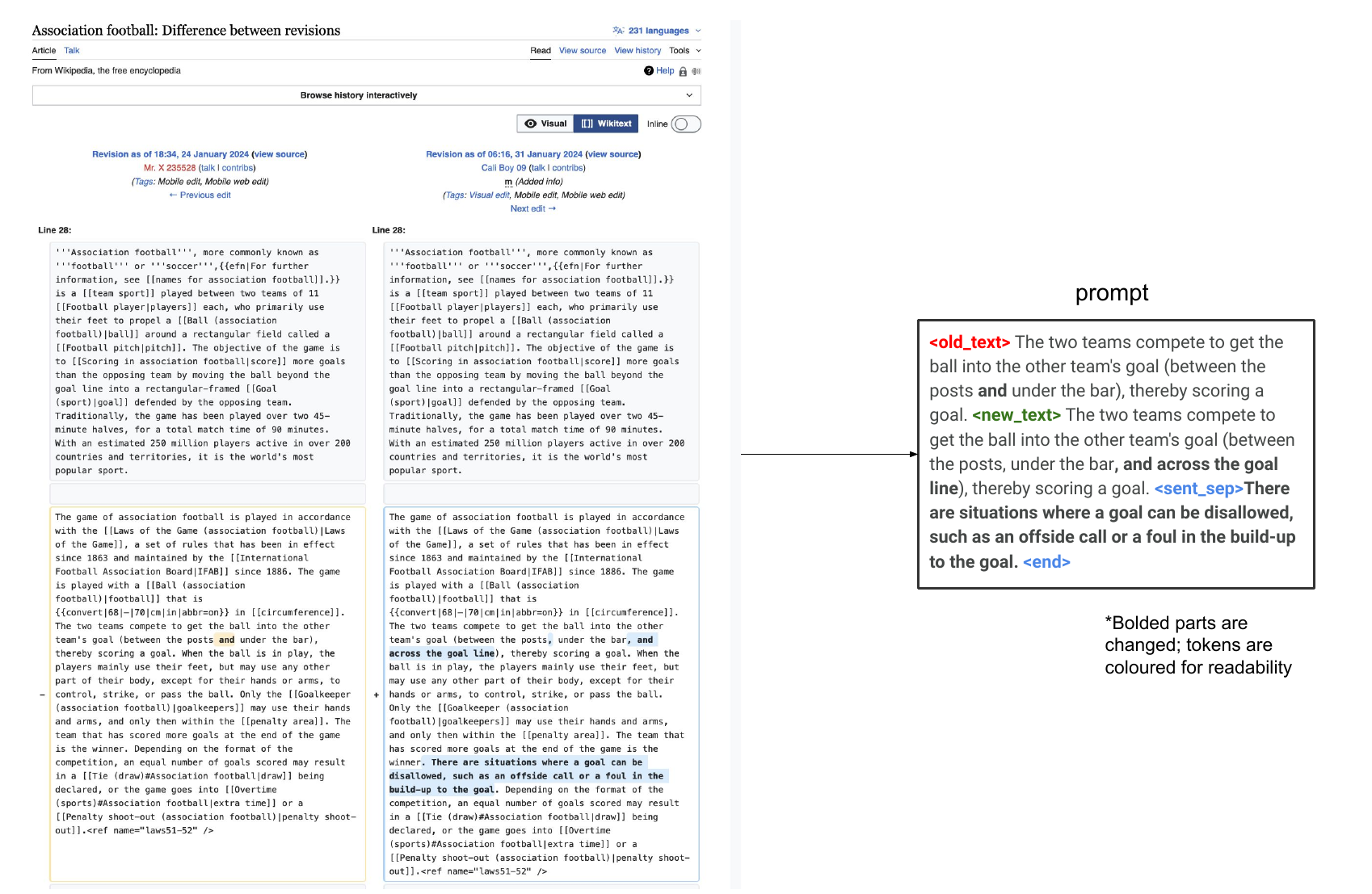}
    \caption{An example of input to the model constructed from an edit diff.}
    \label{fig:input_example}
\end{figure*}

In \Figref{fig:input_example}, we showcase how the input to the model is constructed based on an edit diff.

\section{Implementation hyperparamters and details}
\label{appendix:model_implementation}

We used the long-t5-local-base\footnote{\url{https://huggingface.co/google/long-t5-local-base}} ($\sim$220M parameters) as our base model which we then finetune on the collected training data (see above). The models were trained using the Adam optimizer with learning rate $3 \times 10^{-4}$, 0.1 gradient clipping on the Euclidean norm, and weight decay 0.05. They were trained for 10 epochs with batch size 2 and a polynomial learning-rate scheduler with 1,000 warm-up steps and a final learning rate of $3 \times 10^{-5}$. Training was performed on a single NVIDIA Titan X Maxwell 12GB GPU, taking around 30 hours for each model.

\section{Examples of ground truth and generated edit summaries}
% \usepackage{graphicx}
\begin{table*}[ht]
\centering
\resizebox{\textwidth}{!}{%
\begin{tabular}{|p{0.2\textwidth}|p{0.25\textwidth}|p{0.25\textwidth}|p{0.25\textwidth}|p{0.25\textwidth}|}
\hline
edit                                                 & human editor                                                                                                                                                             & Edisum {[}0\%{]}                                                                                                                                                                         & Edisum {[}100\%{]}                                                                         & GPT-4                                                                                                                                                                                     \\ \hline
\url{https://en.wikipedia.org/w/index.php?diff=1172472704} & added the range                                                                                                                                                          & added location                                                                                                                                                                           & Added location information                                                                 & Added information about the location and altitude of Boveys Lookout and the orientation of the Burton Range.                                                                              \\ \hline
\url{https://en.wikipedia.org/w/index.php?diff=1169223741} & in 1982                                                                                                                                                                  & Added date and citation.                                                                                                                                                                 & Added the year of the speech                                                               & Added the year of Fidel Castro's speech.                                                                                                                                                  \\ \hline
\url{https://en.wikipedia.org/w/index.php?diff=1169701032} & added artist                                                                                                                                                             & Added a wikilink and a link                                                                                                                                                              & Added a new name to the list of artists                                                    & Added Stefan Brüggemann to the list of artists whose works are included.                                                                                                                  \\ \hline
\url{https://en.wikipedia.org/w/index.php?diff=1170976899} & lede; infobox                                                                                                                                                            & infobox (inapplicable); lede; -cats                                                                                                                                                      & Clarified the nature of Denis- Auguste Affre's cause and added information about his title & Updated the description of Denis-Auguste Affre's religious status and clarified the stage of his canonization process.                                                                    \\ \hline
\url{https://en.wikipedia.org/w/index.php?diff=1170699096} & fixed link to aias awards, it was at 5th, not 7th                                                                                                                        & 5th                                                                                                                                                                                      & Corrected the award number from 7th to 5th                                                 & Corrected the year of the Annual Interactive Achievement Awards nomination.                                                                                                               \\ \hline
\url{https://en.wikipedia.org/w/index.php?diff=1169516533} & Amended lists so they are split at 100 m                                                                                                                                 & ""at least", not "at least""                                                                                                                                                             & Clarified the range of heights in the list                                                 & Added information about two different height categories for the ranking of buildings and structures in Greater Manchester.                                                                \\ \hline
\url{https://en.wikipedia.org/w/index.php?diff=1173169263} & UPD Romelu Lukaku                                                                                                                                                        & updated Lukaku                                                                                                                                                                           & Removed the end of the season                                                              & Removed incorrect future date for Romelu Lukaku's tenure at Inter.                                                                                                                        \\ \hline
\url{https://en.wikipedia.org/w/index.php?diff=1171641782} & "This \textbackslash{}"however\textbackslash{}" doesn't make sense here"                                                                                                 & "rmv deprecated, ambiguous "however" per MOS:RELTIME"                                                                                                                                    & Removed unnecessary word                                                                   & Removed the word "However," from the beginning of the sentence.                                                                                                                           \\ \hline
\url{https://en.wikipedia.org/w/index.php?diff=1168487383} & Per feedback given in GA review, elaborated on David King Udall and Ella Stewart Udall in the lead and in Ella Udall's first mention; David Udall was already identified & Added link; improved phrasing and punctuation                                                                                                                                            & Added additional information about the individuals' professions                            & Added professional details for David King Udall, Ella Stewart Udall, and Mary Ann Linton Morgan Udall.                                                                                    \\ \hline
\end{tabular}%
}
\caption{Examples of summaries written by human editors and generated by \model\ and GPT-4.}
\label{tab:summaries_examples}
\end{table*}

In \Tabref{tab:summaries_examples}, we present some of the existing edit summaries, as well as the ones generated with two of our models, \model[100\%] and \model[0\%], and GPT-4.

\section{Human evaluation setup}
\label{appendix:human_eval_setup}

\xhdr{Annotation task}
As mentioned in \Secref{sec:human_eval}, human evaluation was done on 100 samples from the testing dataset, each associated with four edit summaries from different methods, a different Wikipedia edit, and presented with the corresponding web page. The sample size is relatively small as grading these edit summaries is a long and tedious process -- each annotator has to manually assess the edit diff, and sometimes even the whole revisions of the article, in order to understand what was done in it. From the web page, we remove the element showing the actual human edit summary to make sure that the annotators are not aware of the existing edit summary. The web page also shows the ``current'' version of the article, right after the edit, in case the annotators need more context to give their judgement.

Annotators were asked to choose the best and the worst summary out of the four according to the following guidelines:

\noindent \emph{A good edit summary \textbf{should}}:

(1) Summarize what was done in the edit

(2) Cover all the changes performed (either explicitly or by adding something like “and misc”)

(3) Be specific; \eg, a summary “I made some changes” is not specific

(4) Explain why the change was made, if it is unclear from the change itself

% \begin{itemize}
%     \item Summarize what was done in the edit
%     \item Cover all the changes performed (either explicitly or by adding something like “and misc”)
%     \item Be specific; \eg, a summary “I made some changes” is not specific
%     \item Explain why the change was made, if it is unclear from the change itself
% \end{itemize}

\noindent \emph{A good edit summary \textbf{should not}}:

(1) Use uncommon abbreviations

(2) Be too long: it is not supposed to be a paragraph, but a sentence-long summary

(3) Attack other editors' work or be aggressive

% \begin{itemize}
%     \item Use uncommon abbreviations
%     \item Be too long: it is not supposed to be a paragraph, but a sentence-long summary
%     \item Attack other editors' work or be aggressive
% \end{itemize}

\noindent Annotators were provided with examples of good and bad edit summaries, with explanations what makes them good or bad. They were also given Wikipedia's manual of style\footnote{\url{https://en.wikipedia.org/wiki/Wikipedia:Manual_of_Style}} to get familiar with tags that often appear in edit summaries. Finally, to ensure the high-quality of the results, we train them on a few selected samples, teaching them what to look for in the edit summary and making sure they understood the guidelines and the assignment.

\xhdr{Annotators}
%%%%%%% fix
As mentioned in \Secref{sec:human_eval}, we recruit 3 MSc students as annotators for our task. We opt for this option over crowdsourcing platforms for several reasons, all of which ensure high-quality annotations. This task is not trivial for a person not familiar with Wikipedia's norms and rules. For instance, large fraction of edit summaries references Wikipedia's manual of style which might look like irrelevant words to a layperson. By recruiting students, we had more control over the quality of annotators we are taking. All of the annotators were MSc students in computer science, familiar with Wikipedia, but not with our work. On top of that, we had a more straightforward way to train MSc students for this task, as this might be a tricky thing to do with crowdsourcing platforms. Finally, another increasing concern that comes with the use of crowdsourcing platforms is the usage of LLMs by the crowd workers, who today frequently “outsource” text-processing tasks to LLMs to facilitate their work \citep{veselovsky2023artificial}.

The students were paid the equivalent of US\$25 per hour for their work. Conflicts were resolved by one of the authors of the paper.
To measure the agreement between the annotators, we report Kendall rank correlation coefficient (Kendall's $\tau$) between each pair of them. Kendall's $\tau$ is a statistical measure used to assess the degree of association or correlation between two sets of rankings or ordinal data. As our task is a version of a ranking task, we opt to use this metric. For each pair of annotators, and for each sample with four summaries that are rated, we calculate Kendall's $\tau$. To get the value of Kendall's $\tau$ for all 99 samples, we take the average. This way, we report three numbers, a value for each pair of annotators.

% Since this is not a simple task, to ensure high-quality results, instead of relying on the crowdsourcing platforms, we recruited 3 MSc students to perform the annotation. The students were unaware of the topic of the paper to avoid bias. They were paid the equivalent of US\$25 per hour for their work. Conflicts were resolved by one of the authors of the paper.
% To measure the agreement between the annotators, we report Kendall rank correlation coefficient (Kendall's $\tau$) between each pair of them. Kendall's $\tau$ is a statistical measure used to assess the degree of association or correlation between two sets of rankings or ordinal data. Since our annotation task can be seen as a ranking task, we choose this as a suitable measure. For each pair of annotators, and for each sample with four summaries that are rated, we calculate Kendall's $\tau$. To get the value of Kendall's $\tau$ for all 99 samples, we take the average. This way, we report three numbers, a value for each pair of annotators.

\section{Plackett-Luce model for ranking our methods}
\label{appendix:plackett-luce}
In \Tabref{tab:plackett_luce}, we present the obtained result of fitting a Plackett-Luce model to our ranking data from the human evaluation. Results show that \model[100\%] performs similarly to the human editors, and even has a slight advantage over them. GPT-4 performs the best, while \model[0\%] performs the worst. These are in line with our results from human evaluation.

\begin{table}[h]
\centering
\resizebox{\columnwidth}{!}{
\setlength{\tabcolsep}{5pt}
\begin{tabular}{@{}l|cccc@{}}
\toprule
\textbf{Model} & \model[0\%] & \model[100\%] & human editors & GPT-4 \\
\midrule
\textbf{Parameter} & 0.072 \scriptsize± 0.022 & 0.308 \scriptsize± 0.019 & 0.276 \scriptsize± 0.027 & 0.346 \scriptsize± 0.023  \\
\bottomrule
\end{tabular}
}
\caption{Parameters obtained for each method with Plackett-Luce model.}
\label{tab:plackett_luce}
\end{table}

\section{Annotation procedure for error analysis}
\label{appendix:error_annotation}

\xhdr{Data}
We annotate in total 150 samples, 50 for each method (GPT-4, \model[100\%] and human editors). For each method, samples were chosen to have as similar as possible fractions of samples that won, lost or were not chosen as either. For GPT-4, there was only 1 sample chosen as the worst, so we do not include this category in the analysis.

\xhdr{Annotation procedure}
Samples were annotated according to two meta-categories: ``what'' (content of the edit) and ``why'' (reason for the edit). Taxonomies for each of the meta-categories ware derived after manual inspection of the subset of the outputs. For ``why'', we settle on three simple categories: missing, correct, and incorrect. For ``what'', we derived the following categories:
\begin{enumerate}
    \item Correct: edit summary is fully correct and exhaustive
    \item No change: the summary indicates that no change was performed
    \item Not specific: the summary is not describing exact changes that were performed
    \item Unclear: the summary seems to be pointing to the correct modifications, but is hard to understand without looking at the diff
    \item Unexhaustive: the summary does not cover all changes performed
    \item Unrelated: the summary describes unrelated edit
\end{enumerate}

Annotation was done by one of the paper authors. The annotator chose one of the categories from the taxonomy for each of the presented edit summaries. To confirm the validity of the results, another author annotated 30 random samples. We calculated the agreement between the two annotators on those 30 samples using Cohen's kappa. For ``what'' meta-category, Cohen's kappa is 0.60, while for ``why'' it is 0.67. Both of these numbers indicate high overlap between the annotators.

\section{Additional automatic evaluation}
\label{appendix:automatic_eval}

In \Tabref{tab:rouge_bert_score}, we present ROUGE and BERT score for each evaluated model from \Secref{sec:auto_eval_res}. Results are mostly in line with MoverScore from the same section, confirming the superiority of GPT-4 for this task.

\begin{table}
\centering
\resizebox{\columnwidth}{!}{
\setlength{\tabcolsep}{3pt}
\begin{tabular}{c|cccc}
\toprule
\textbf{Method} & \textbf{BERT score} & \textbf{ROUGE-1} & \textbf{ROUGE-2} & \textbf{ROUGE-L} \\
\midrule
Edisum[0\%] & 0.803 \scriptsize± 0.006 & 0.077 \scriptsize± 0.022 & 0.026 \scriptsize± 0.016 & 0.076 \scriptsize± 0.020 \\
Edisum[25\%] & 0.823 \scriptsize± 0.006 & 0.101 \scriptsize± 0.025 & 0.026 \scriptsize± 0.014 & 0.076 \scriptsize± 0.020 \\
Edisum[50\%] & 0.820 \scriptsize± 0.007 & 0.092 \scriptsize± 0.020 & 0.020 \scriptsize± 0.013 & 0.087 \scriptsize± 0.019 \\
Edisum[75\%] & 0.833 \scriptsize± 0.005 & 0.094 \scriptsize± 0.021 & 0.015 \scriptsize± 0.009 & 0.087 \scriptsize± 0.017 \\
Edisum[100\%] & 0.833 \scriptsize± 0.004 & 0.090 \scriptsize± 0.017 & 0.012 \scriptsize± 0.007 & 0.083 \scriptsize± 0.017\\
\midrule
GPT-3.5 & 0.836 \scriptsize± 0.004 & 0.100 \scriptsize± 0.017 & 0.017 \scriptsize± 0.009 & 0.095 \scriptsize± 0.015 \\
GPT-4 & 0.837 \scriptsize± 0.004 & 0.118 \scriptsize± 0.016 & 0.025 \scriptsize± 0.010 & 0.110 \scriptsize± 0.016 \\
Llama-3-8B & 0.637 \scriptsize± 0.045 & 0.031 \scriptsize± 0.010 & 0.003 \scriptsize± 0.003 & 0.029 \scriptsize± 0.011 \\
\bottomrule
\end{tabular}
}
\caption{Automatic evaluation with ROUGE and BERT score.}
\label{tab:rouge_bert_score}
\end{table}

% \section{Ablation study on size of \model}
% \label{appendix:ablation_size}

% \marija{TODO}

\end{document}